\documentclass[11pt,letterpaper]{article}
\usepackage{emnlp2017}
\usepackage{times}
\usepackage{latexsym}
\usepackage{amsmath}
\usepackage{amsfonts}
\usepackage{multirow}
\usepackage{graphicx}

\emnlpfinalcopy

\def\emnlppaperid{***}

\def\model/{{\sc KeyVec}}
\def\docrepr/{\mathbf{d}}

\title{KeyVec: Key-semantics Preserving Document Representations}

\author{Bin Bi \and Hao Ma \\
  Microsoft Research\\
  One Microsoft Way\\
  Redmond, WA 98052\\
  {\tt \{bibi, haoma\}@microsoft.com}}

\begin{document}

\maketitle

\begin{abstract}
Previous studies have demonstrated the empirical success of word embeddings in various applications. In this paper, we investigate the problem of learning distributed representations for text documents which many machine learning algorithms take as input for a number of NLP tasks.

We propose a neural network model, \model/, which learns document representations with the goal of preserving key semantics of the input text. It enables the learned low-dimensional vectors to retain the topics and important information from the documents that will flow to downstream tasks. Our empirical evaluations show the superior quality of \model/ representations in two different document understanding tasks.
\end{abstract}

\section{Introduction}

In recent years, the use of word representations, such as {\tt word2vec}~\cite{Mikolov:2013,Mikolov:2013nips} and {\tt GloVe}~\cite{Pennington:2014}, has become a key ``secret sauce'' for the success of many natural language processing (NLP), information retrieval (IR) and machine learning (ML) tasks. The empirical success of word embeddings raises an interesting research question: \emph{Beyond words, can we learn fixed-length distributed representations for pieces of texts?} The texts can be of variable-length, ranging from paragraphs to documents. Such document representations play a vital role in a large number of downstream NLP/IR/ML applications, such as text clustering, sentiment analysis, and document retrieval, which treat each piece of text as an instance. Learning a good representation that captures the semantics of each document is thus essential for the success of such applications.

In this paper, we introduce \emph{\model/}, a neural network model that learns densely distributed representations for documents of variable-length. In order to capture semantics, the document representations are trained and optimized in a way to recover key information of the documents. In particular, given a document, the \model/ model constructs a fixed-length vector to be able to predict both salient sentences and key words in the document. In this way, \model/ conquers the problem of prior embedding models which treat every word and every sentence equally, failing to identify the key information that a document conveys. As a result, the vectorial representations generated by \model/ can naturally capture the topics of the documents, and thus should yield good performance in downstream tasks.

We evaluate our \model/ on two text understanding tasks: document retrieval and document clustering. As shown in the experimental section~\ref{sec:exp}, \model/ yields generic document representations that perform better than state-of-the-art embedding models.

\begin{figure*}[h]
\centering
\includegraphics[width=0.7\linewidth]{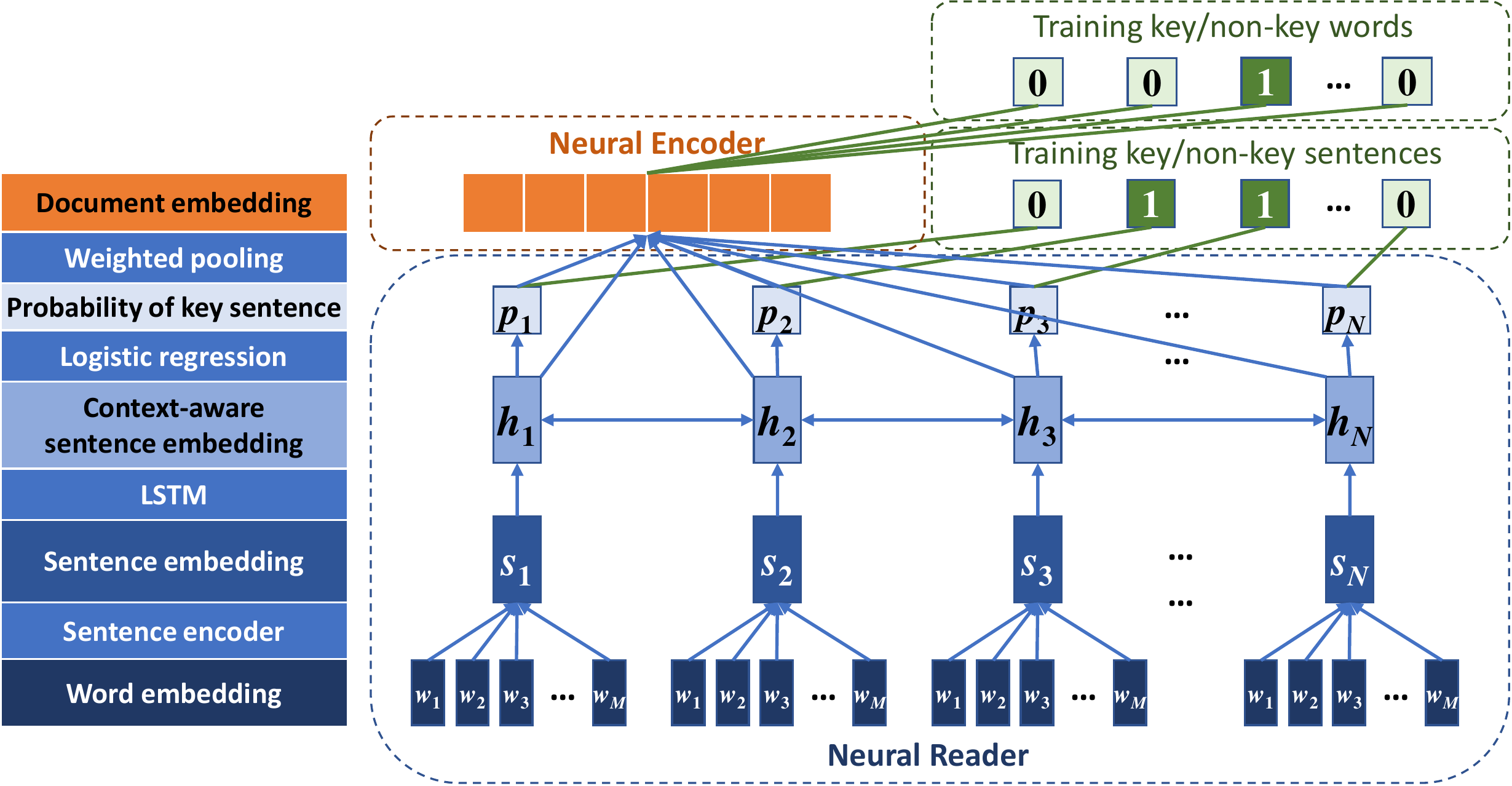}
\vspace{-5pt}
\caption{\model/ Model (\emph{best viewed in color})}
\label{fig:model}
\vspace{-10pt}
\end{figure*}

\section{Related Work}
\vspace{-5pt}
Le~\emph{et al.} proposed a \emph{Paragraph Vector} model, which extends {\tt word2vec} to vectorial representations for text paragraphs~\cite{Le:2014,Dai:2015}. It projects both words and paragraphs into a single vector space by appending paragraph-specific vectors to typical {\tt word2vec}. Different from our \model/, Paragraph Vector does not specifically model key information of a given piece of text, while capturing its sequential information. In addition, Paragraph Vector requires extra iterative inference to generate embeddings for unseen paragraphs, whereas our \model/ embeds new documents simply via a single feed-forward run.

In another recent work~\cite{Djuric:2015}, Djuric~\emph{et al.} introduced a Hierarchical Document Vector (HDV) model to learn representations from a document stream. Our \model/ differs from HDV in that we do not assume the existence of a document stream and HDV does not model sentences.

\section{\model/ Model}

Given a document $D$ consisting of $N$ sentences $\{ s_1, s_2, \ldots, s_N \}$, our \model/ model aims to learn a fixed-length vectorial representation of $D$, denoted as $\docrepr/$. Figure~\ref{fig:model} illustrates an overview of the \model/ model consisting of two cascaded neural network components: a {\bf Neural Reader} and a {\bf Neural Encoder}, as described below.

\subsection{Neural Reader}

The Neural Reader learns to understand the topics of every given input document with paying attention to the salient sentences. It computes a dense representation for each sentence in the given document, and derives its probability of being a salient sentence. The identified set of salient sentences, together with the derived probabilities, will be used by the Neural Encoder to generate a document-level embedding.

Since the Reader operates in embedding space, we first represent discrete words in each sentence by their \textit{word embeddings}. The sentence encoder in Reader then derives \textit{sentence embeddings} from the word representations to capture the semantics of each sentence. After that, a Recurrent Neural Network (RNN) is employed to derive document-level semantics by consolidating constituent sentence embeddings. Finally, we identify key sentences in every document by computing the probability of each sentence being salient.

\subsubsection{Sentence Encoder}
Specifically, for the $i$-th sentence $s_i=\{ w_{i1}, w_{i2}, \ldots, w_{iM} \}$ with $M$ words, Neural Reader maps each word $w_{im}$ into a word embedding $\mathbf{w}_{im} \in \mathbb{R}^{d_{\mathcal{W}}}$. Pre-trained word embeddings like {\tt word2vec} or {\tt GloVe} may be used to initialize the embedding table. In our experiments, we use domain-specific word embeddings trained by {\tt word2vec} on our corpus.

Given the set of word embeddings for each sentence, Neural Reader then derives sentence-level embeddings $\mathbf{s_i}$ using a sentence encoder $g(\cdot)$:
\vspace{-5pt}
\begin{equation}
\vspace{-5pt}
    \mathbf{s}_i = g(\mathbf{w}_{i1}, \mathbf{w}_{i2}, \ldots, \mathbf{w}_{iM}),
\end{equation}
where $g(\cdot)$ is implemented by a Convolutional Neural Network (CNN) with a max-pooling operation, in a way similar to~\cite{Kim:2014}. Note that other modeling choices, such as an RNN, are possible as well. We used a CNN here because of its simplicity and high efficiency when running on GPUs. The sentence encoder generates an embedding $\mathbf{s}_i$ of 150 dimensions for each sentence.

\subsubsection{Identifying Salient Sentences}
Given the embeddings of sentences $\{ \mathbf{s}_1, \mathbf{s}_2, \ldots, \mathbf{s}_N \}$ in a document $d$, Neural Reader computes the probability of each sentence $s_i$ being a key sentence, denoted as $p(s_i|d)$.

We employ a Long Short-Term Memory (LSTM)~\cite{Hochreiter:1997} to compose constituent sentence embeddings into a document representation. At the $i$-th time step, LSTM takes as input the current sentence embedding $\mathbf{s}_i$, and computes a hidden state $\mathbf{h}_i$. We place an LSTM in both directions, and concatenate the outputs of the two LSTMs. For the $i$-th sentence, $\mathbf{h}_i$ is semantically richer than sentence embedding $\mathbf{s}_i$, as $\mathbf{h}_i$ incorporates the context information from surrounding sentences to model the temporal interactions between sentences. The probability of sentence $s_i$ being a key sentence then follows a logistic sigmoid of a linear function of $\mathbf{h}_i$:
\vspace{-5pt}
\begin{equation}
    p(s_i|d) = \sigma (\mathbf{w}_l^\top\mathbf{h}_i+b_l),
\vspace{-5pt}
\end{equation}
where $\mathbf{w}_l\in \mathbb{R}^{|\mathbf{h}_i|}$ is a trainable weight vector, and $b_l\in \mathbb{R}$ is a trainable bias scalar.

\subsection{Neural Encoder}
The Neural Encoder computes document-level embeddings based on the salient sentences identified by the Reader. In order to capture the topics of a document and the importance of its individual sentences, we perform a weighted pooling over the constituent sentences, with the weights specified by $p(s_i|d)$, which gives the document-level embedding $\mathbf{d}$ through a $tanh$ transformation:
\vspace{-5pt}
\begin{equation}
    \mathbf{d} = \tanh (\mathbf{W}_d\left(\sum_{i=1}^N \frac{p(s_i|d)}{\sum_{j=1}^N p(s_j|d)} \cdot \mathbf{h}_i\right)+\mathbf{b}_d),
\end{equation}
where $\mathbf{W}_d\in \mathbb{R}^{|\mathbf{d}|\times |\mathbf{h}|}$ is a trainable weight matrix, and $\mathbf{b}_d\in \mathbb{R}^{|\mathbf{d}|}$ is a trainable bias vector.

Weighted pooling functions are commonly used as the attention mechanism~\cite{Bahdanau:2015} in neural sequence learning tasks. The ``share'' each sentence contributes to the final embedding is proportional to its probability of being a salient sentence. As a result, $\mathbf{d}$ will be dominated by salient sentences with high $p(s_i|d)$, which preserves the key information in a document, and thus allows long documents to be encoded and embedded semantically.

\section{Model Learning}
\vspace{-5pt}
In this section, we describe the learning process of the parameters of \model/. Similarly to most neural network models, \model/ can be trained using Stochastic Gradient Descent (SGD), where the Neural Reader and Neural Encoder are jointly optimized. In particular, the parameters of Reader and Encoder are learned simultaneously by maximizing the joint likelihood of the two components:
\vspace{-5pt}
\begin{equation}
\vspace{-5pt}
    \mathcal{L} = \mathcal{L}_{\tt read} + \mathcal{L}_{\tt enc},
    \label{joint_obj}
\end{equation}
where $\mathcal{L}_{\tt read}$ and $\mathcal{L}_{\tt enc}$ denotes the log likelihood functions of Reader and Encoder, respectively.

\subsection{Reader's Objective: $\mathcal{L}_{\tt read}$}
To optimize Reader, we take a surrogate approach to heuristically generate a set of salient sentences from a document collection, which constitute a training dataset for learning the probabilities of salient sentences $p(s_i|d, \theta)$ parametrized by $\theta$. More specifically, given a training set $\mathcal{D}$ of documents (e.g., body-text of research papers) and their associated summaries (e.g., abstracts) $\{\langle d_k, S_k \rangle\}_{k=1}^{|\mathcal{D}|}$, where $S_k$ is a gold summary of document $d_k$, we employ a state-of-the-art sentence similarity model, DSSM~\cite{Huang:2013,Shen:2014}, to find the set of top-$K$\footnote{$K=10$ in our experiments} sentences $S^*_k=\{ s'_i \}$ in $d_k$, such that the similarity between $s'_i \in S^*_k$ and any sentence in the gold summary $S_k$ is above a pre-defined threshold. Note that here we assume each training document is associated with a gold summary composed of sentences that might not come from $d_k$. We make this assumption only for the sake of generating the set of salient sentences $S^*_k$ which is usually not readily available.

The log likelihood objective of the Neural Reader is then given by maximizing the probability of $S^*_k$ being the set of key sentences, denoted as $p(S^*_k|d_k)$:
\vspace{-5pt}
\begin{equation}
\vspace{-5pt}
\begin{split}
&\mathcal{L}_{\tt read}= \sum_{k=1}^{|\mathcal{D}|} \log p(S^*_k|d_k) \\
&= \sum_{k=1}^{|\mathcal{D}|} \log \Big( \prod_{s_i \in S^*_k} p(s_i|d_k) \cdot \prod_{s_i \in d_k\backslash S^*_k} \big(1 - p(s_i|d_k) \big) \Big) \\
&= \sum_{k=1}^{|\mathcal{D}|} \Big( \sum_{s_i \in S^*_k} \log p(s_i|d_k)+\sum_{s_i \in d_k\backslash S^*_k} \log \big(1 - p(s_i|d_k) \big) \Big),
\end{split}
\label{eq:mlt_sum}
\end{equation}
where $d_k\backslash S^*_k$ is the set of non-key sentences. Intuitively, this likelihood function gives the probability of each sentence in the generated key sentence set $S^*_k$ being a key sentence, and the rest of sentences being non-key ones.

\subsection{Encoder's Objective: $\mathcal{L}_{\tt enc}$}

The final output of Encoder is a document embedding $\mathbf{d}$, derived from LSTM's hidden states $\{ \mathbf{h} \}$ of Reader. Given our goal of developing a general-purpose model for embedding documents, we would like $\mathbf{d}$ to be semantically rich to encode as much key information as possible. To this end, we impose an additional objective on Encoder: the final document embedding needs to be able to reproduce the key words in the document, as illustrated in Figure~\ref{fig:model}.

In document $d_k$, the set of key words $W_k$ is composed of top 30 words in $S_k$ (i.e., the gold summary of $d_k$) with the highest TF-IDF scores. Encoder's objective is then formalized by maximizing the probability of predicting the key words in $W_k$ using the document embedding $\mathbf{d}_k$:
\vspace{-5pt}
\begin{equation}
\vspace{-5pt}
    \mathcal{L}_{\tt enc} = \sum_{k=1}^{|\mathcal{D}|} \sum_{w \in W_k} \log p(w \in W_k|\mathbf{d}_k),
    \label{eq:mlt_enc}
\end{equation}
where $p(w \in W_k|\mathbf{d}_k)$ is implemented as a softmax function with output dimensionality being the size of the vocabulary.

Combining the objectives of Reader and Encoder yields the joint objective function in Eq~\eqref{joint_obj}. By jointly optimizing the two objectives with SGD, the \model/ model is capable of learning to identify salient sentences from input documents, and thus generating semantically rich document-level embeddings.

\begin{table}
\centering
\small{
\begin{tabular}{|c|c|c|c|}
\hline
Model & P@10 & MAP & MRR\\
\hline\hline
word2vec averaging & \multirow{2}{*}{0.221} & \multirow{2}{*}{0.176} & \multirow{2}{*}{0.500}\\
 (public release 300d) & & &\\
\hline
word2vec averaging & \multirow{2}{*}{0.223} & \multirow{2}{*}{0.193} & \multirow{2}{*}{0.546}\\
(academic corpus) & & &\\
\hline
Paragraph Vector & 0.227 & 0.177 & 0.495\\
\hline
\model/ & \textbf{0.279} & \textbf{0.232} & \textbf{0.619}\\
\hline
\end{tabular}
}
\caption{\emph{Evaluation of document retrieval with different embedding models}}
\label{table:retrieval}
\vspace{-10pt}
\end{table}

\section{Experiments and Results}
\vspace{-5pt}
\label{sec:exp}
To verify the effectiveness, we evaluate the \model/ model on two text understanding tasks that take continuous distributed vectors as the representations for documents: \emph{document retrieval} and \emph{document clustering}.

\subsection{Document Retrieval}
The goal of the document retrieval task is to decide if a document should be retrieved given a query. In the experiments, our document pool contained 669 academic papers published by IEEE, from which top-$k$ relevant papers are retrieved. We created 70 search queries, each composed of the text in a Wikipedia page on a field of study (e.g., \url{https://en.wikipedia.org/wiki/Deep_learning}). We retrieved relevant papers based on cosine similarity between document embeddings of 100 dimensions for Wikipedia pages and academic papers. For each query, a good document-embedding model should lead to a list of academic papers in one of the 70 fields of study.


Table~\ref{table:retrieval} presents \emph{P@10}, \emph{MAP} and \emph{MRR} results of our \model/ model and competing embedding methods in academic paper retrieval. \emph{word2vec averaging} generates an embedding for a document by averaging the {\tt word2vec} vectors of its constituent words. In the experiment, we used two different versions of {\tt word2vec}: one from public release, and the other one trained specifically on our own academic corpus (113 GB). From Table~\ref{table:retrieval}, we observe that as a document-embedding model, Paragraph Vector gave better retrieval results than \emph{word2vec averagings} did. In contrast, our \model/ outperforms all the competitors given its unique capability of capturing and embedding the key information of documents.

\subsection{Document Clustering}
In the document clustering task, we aim to cluster the academic papers by the venues in which they are published. There are a total of 850 academic papers, and 186 associated venues which are used as ground-truth for evaluation. Each academic paper is represented as a vector of 100 dimensions.

\begin{table}
\centering
\small{
\begin{tabular}{|c|c|c|c|}
\hline
Model & F1 & V-measure & ARI\\
\hline\hline
word2vec averaging & \multirow{2}{*}{0.019} & \multirow{2}{*}{0.271} & \multirow{2}{*}{0.003}\\
 (public release 300d) & & &\\
\hline
word2vec averaging & \multirow{2}{*}{0.079} & \multirow{2}{*}{0.548} & \multirow{2}{*}{0.066}\\
(academic corpus) & & &\\
\hline
Paragraph Vector & 0.083 & 0.553 & 0.070\\
\hline
\model/ & \textbf{0.090} & \textbf{0.597} & \textbf{0.079}\\
\hline
\end{tabular}
}
\caption{\emph{Evaluation of document clustering with different embedding models}}
\label{table:clustering}
\vspace{-10pt}
\end{table}

To compare embedding methods in academic paper clustering, we calculate \emph{F1}, \emph{V-measure} (a conditional entropy-based clustering measure~\cite{Rosenberg:2007}), and \emph{ARI} (Adjusted Rand index~\cite{Hubert:1985}). As shown in Table~\ref{table:clustering}, similarly to document retrieval, Paragraph Vector performed better than \emph{word2vec averagings} in clustering documents, while our \model/ consistently performed the best among all the compared methods.


\section{Conclusions}

In this work, we present a neural network model, \model/, that learns continuous representations for text documents in which key semantic patterns are retained.

In the future, we plan to employ the Minimum Risk Training scheme to train Neural Reader directly on original summary, without needing to resort to a sentence similarity model.

\bibliography{keyvec}
\bibliographystyle{emnlp_natbib}

\end{document}